\documentclass[10pt,twocolumn,letterpaper]{article}

\usepackage{iccv}
\usepackage{pgfplots}
\usepackage{pgf,tikz}
\usepackage{times}
\usepackage{epsfig}
\usepackage{graphicx}
\usepackage{amsmath}
\usepackage{amssymb}
\usepackage{amsthm}
\usepackage{overpic}
\usepackage{wrapfig}
\usepackage{bm}

\newcommand{\X}{\mathcal{X}}
\newcommand{\Y}{\mathcal{Y}}
\newcommand{\Z}{\mathcal{Z}}
\newcommand{\C}{\mathbf{C}}

\newcommand{\dx}{\mathrm{d}\mu}
\newcommand{\bb}[1]{\boldsymbol{\mathrm{#1}}}

\graphicspath{ {figures/} }
\usepackage[pagebackref=true,breaklinks=true,letterpaper=true,colorlinks,bookmarks=false]{hyperref}

 \iccvfinalcopy % *** Uncomment this line for the final submission

\def\iccvPaperID{2649} % *** Enter the ICCV Paper ID here

\ificcvfinal\fi
\begin{document}

%%%%%%%%% TITLE
\title{Deep Functional Maps:\\Structured Prediction for Dense Shape Correspondence}

\author{
Or Litany$^{1,2}$
\hspace{1cm}
Tal Remez$^{1}$
\\
Emanuele Rodol\`a$^{3,4}$
\hspace{1cm}
Alex Bronstein$^{2,5}$
\hspace{1cm}
Michael Bronstein$^{2,3}$
\vspace{0.2cm}\\
{
$^1$Tel Aviv University \hspace{1mm}
$^2$Intel \hspace{1mm}
$^3$USI Lugano \hspace{1mm}
$^4$Sapienza University of Rome \hspace{1mm}
$^5$Technion
}
}

\maketitle
%\thispagestyle{empty}

%%%%%%%%% ABSTRACT
\begin{abstract}
We introduce a new framework for learning dense correspondence between deformable 3D shapes. Existing learning based approaches model shape correspondence as a labelling problem, where each point of a query shape receives a label identifying a point on some reference domain; the correspondence is then constructed a posteriori by composing the label predictions of two input shapes. We propose a paradigm shift and design a structured prediction model in the space of functional maps, linear operators that provide a compact representation of the correspondence. We model the learning process via a deep residual network which takes dense descriptor fields defined on two shapes as input, and outputs a soft map between the two given objects. The resulting correspondence is shown to be accurate on several challenging benchmarks comprising multiple categories, synthetic models, real scans with acquisition artifacts, topological noise, and partiality.

%We introduce a new framework for learning dense correspondences between deformable 3D shapes. Differently from existing approaches, our method operates a structured prediction in the space of functional maps, linear operators that provide a compact representation of the correspondence. We model the learning process via a deep residual network which takes dense descriptor fields defined on two shapes as input, and outputs a functional map between the two given objects. The resulting correspondence is highly accurate, and significantly outperforms existing methods on real data according to the most recent benchmarks.
\end{abstract}

%%%%%%%%% BODY TEXT
\section{Introduction}
3D acquisition technology has made great progress in the last decade, and is being rapidly incorporated into commercial products ranging from Microsoft Kinect \cite{zhang2012microsoft} for gaming, to LIDARs used in autonomous cars. An essential building block for application design in many of these domains is to recover 3D shape correspondences in a fast and reliable way. While handling real-world scanning artifacts is a challenge by itself, additional complications arise from non-rigid motions of the objects of interest (typically humans or animals). Most non-rigid shape correspondence methods employ local descriptors that are designed to achieve robustness to noise and deformations; however, relying on such ``handcrafted'' descriptors can often lead to inaccurate solutions in practical settings. 
%
%A popular family of methods that have shown to produce smooth and accurate correspondence from a small set of corresponding functions is called functional maps \cite{ovsjanikov12}. Specifically, the correspondence is modeled as a linear map between spaces of functions, and can be recovered provided a e.g. pointwise descriptors.
%
Partial remedy to this was brought by the recent line of works on learning shape correspondence \cite{litman2014learning,rodola14,masci15,WFT2015,add16,boscaini2016learning,monet}. 
%
%These approaches optimize descriptor similarity for corresponding point while distinguishing between non-corresponding ones. 
%
A key drawback of these methods lies in their emphasis on learning a descriptor that would help in identifying corresponding points, or on learning a labeling with respect to some reference domain. On the one hand, by focusing on the descriptor, the learning process remains agnostic to the way the final correspondence is computed, and costly post-processing steps are often necessary in order to obtain accurate solutions from the learned descriptors. On the other hand, methods based on a label space are restricted to a fixed number of points and rely on the adoption of an intermediate reference model.

\begin{figure}[t]
  \centering
\begin{overpic}
[trim=0cm 0cm 0cm 0cm,clip,width=1.0\linewidth]{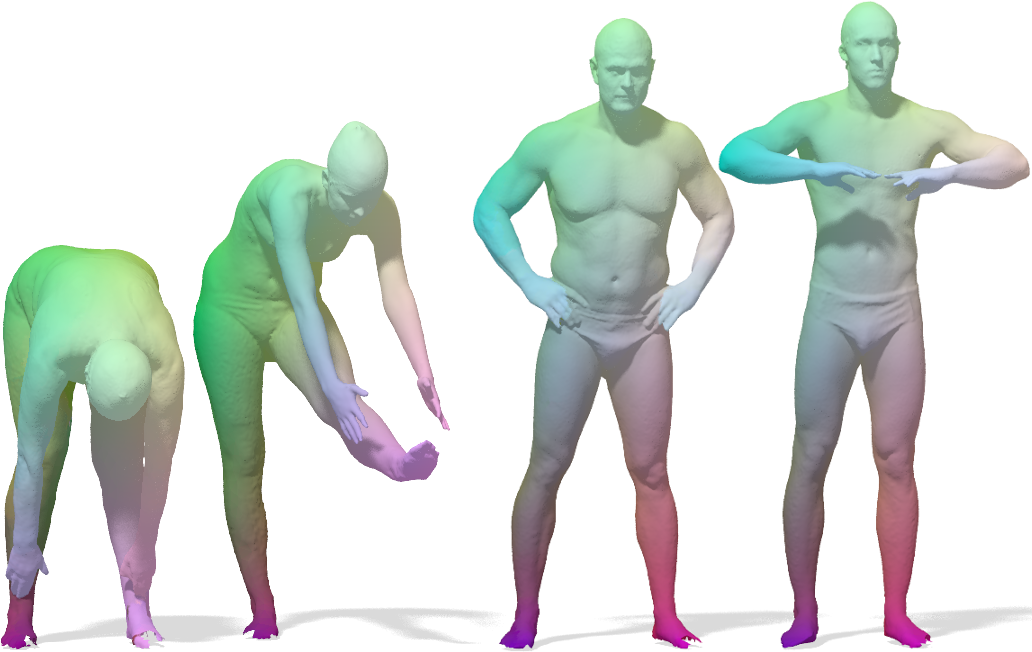}
%\put(2,-1){\footnotesize source}
\end{overpic}
  \caption{\label{fig:teaser}Correspondence results obtained by our network model on two pairs from the FAUST real scans challenge. Corresponding points are assigned the same color. The average error for the left and right pairs is $5.21$cm and $2.34$cm respectively. Accurate correspondence is obtained despite mesh ``gluing'' in areas of contact.}
\end{figure}

\paragraph*{Contribution.}
In this work we propose a {\em task-driven} approach for descriptor learning, by including the computation of the correspondence directly as part of the learning procedure. Perfect candidates for this task are neural networks, 
%
%which have been proven to have a huge success in computer vision tasks in general and shape matching in particular, and
%
 due to their inherent flexibility to the addition of computational blocks. 
%
%Indeed, we show significant improvement in accuracy with respect to other learning based methods on a few recent benchmarks.
%
Our main contributions can be summarized as follows:
\begin{itemize}
	\item We introduce a new {\em structured prediction} model for shape correspondence. Our framework allows end-to-end training: it takes base descriptors as input, and returns matches.
	\item We show that our approach consistently outperforms existing descriptor and correspondence learning methods on several recent benchmarks.
\end{itemize}

\section{Related work}
\vspace{1ex}\noindent\textbf{Shape correspondence } is an active area of research in computer vision, graphics, and pattern recognition, with a variety of both classical and recent methods \cite{bronstein2006generalized,kim11,chen15}. Since a detailed review of the literature would be out of scope for this paper, we refer the interested reader to the recent surveys on shape correspondence  \cite{van2011survey,biasotti2015recent}. 
More closely related to our approach is the family of methods based on the notion of {\em functional maps} \cite{ovsjanikov12}, modeling correspondences as linear operators between spaces of functions on manifolds. In the truncated Laplacian eigenbases, such operators can be compactly encoded as small matrices -- drastically reducing the amount of variables to optimize for, and leading to an increased flexibility in manipulating correspondences. This representation has been adopted and extended in several follow-up works \cite{pokrass13,huangn14,kovnatsky15,aflalo2016spectral,eynard2016coupled,rodola16-partial,litany16,litany17fully,dorian17}, demonstrating state-of-the-art performance in multiple settings.

While being often adopted as a useful tool for the post-processing of some initial correspondence, functional maps have rarely been employed as a building block in correspondence learning pipelines.

%Classical non-rigid correspondence methods for approximately isometric or topologically equivalent shapes include \cite{bronstein2006generalized, Lipman09,kim11}. These methods build on the existance of good descriptors \td{ema: is this true?}. 

\vspace{1ex}\noindent\textbf{Descriptor learning.}
Descriptor and feature learning are key topics in computer vision, and in recent years they have been actively investigated by the shape analysis community.
%Many such descriptors have been previosely proposed, including the extension of classical rigid descriptors by replacing the Euclidean metric with its geodesic counterpart \cite{hamza2003geodesic, elad2003bending}. 
%
%Intrinsic methods based on the eigenfunctions and the eigenvalues of the Laplacian include HKS \cite{sun09} and the physically based WKS \cite{aubry2011wave}. A popular descriptor \td{...what can we say here?} is SHOT \cite{tombari10}.  \td{shape context [6] and spin image [7] descriptors, as integral volume descriptors,and multiscale local features [9]}. Improving upon those hand-crafted descriptors, clasical machine learning techniques were used to produce data-driven descriptors.
%
Litman~\etal~\cite{litman2014learning} introduced {\em optimal spectral descriptors}, a data-driven parametrization of the spectral descriptor model (\ie, based on the eigen-decomposition of the Laplacian), demonstrating noticeable improvement upon the classical axiomatic constructions \cite{sun09,aubry2011wave}. A similar approach was subsequently taken in \cite{windheuser2014optimal} and more recently in \cite{WFT2015,add16,clutter}. 
Perhaps most closely related to ours is the approach of Corman \etal \cite{corman2014supervised}, where combination weights for an input set of descriptors are learned in a supervised manner. Similarly to our approach, they base their construction upon the functional map framework \cite{ovsjanikov12}. While their optimality criterion is defined in terms of deviation from a ground-truth functional map in the {\em spectral} domain, we aim at recovering an optimal map in the {\em spatial} domain. Our structured prediction model will be designed with this requirement in mind.

\vspace{1ex}\noindent\textbf{Correspondence learning.}
Probably the first example of learning correspondence for deformable 3D shapes is the ``shallow'' random forest approach of Rodol{\`a}~\etal~\cite{rodola14}. More recently, Wei~\etal\cite{wei2016dense} employed a classical (extrinsic) CNN architecture trained on huge training sets for learning invariance to pose changes and clothing. Convolutional neural networks on non-Euclidean domains (surfaces) were first considered by Masci~\etal\cite{masci15} with the introduction of the geodesic CNN model, a deep learning architecture where the classical convolution operation is replaced by an intrinsic (albeit, non-shift invariant) counterpart. The framework was shown to produce promising results in descriptor learning and shape matching applications, and was recently improved by Boscaini~\etal\cite{boscaini2016learning} and generalized further by Monti~\etal~\cite{monet}. 
These methods are instances of a broader recent trend of {\em geometric deep learning} attempting to generalize successful deep learning paradigms to data with non-Euclidean underlying structure such as manifolds or graphs \cite{gdl}.

\section{Background}\label{sec:bg}
%
%----------------------------------------------------------------
\vspace{1ex}\noindent\textbf{Manifolds.}
We model shapes as two-dimensional Riemannian manifolds $\X$ (possibly with boundary $\partial\X$) equipped with the standard measure $\dx$ induced by the volume form. Throughout the paper we will consider the space of functions $L^2(\X)=\{f:\X\to\mathbb{R}~|~\langle f,f\rangle_\X<\infty\}$, with the standard manifold inner product $\langle f,g\rangle_\X=\int_\X f\cdot g~\dx$.

The positive semi-definite Laplace-Beltrami operator $\Delta_\X$ generalizes the notion of Laplacian from Euclidean spaces to surfaces. It admits an eigen-decomposition ${\Delta_\X \phi_i = \lambda_i \phi_i}$ (with proper boundary conditions if ${\partial\X\neq\emptyset}$), where the eigenvalues form a discrete spectrum $0=\lambda_1\le\lambda_2\le\dots$ and the eigenfunctions $\phi_1,\phi_2,\dots$ form an orthonormal basis for $L^2(\X)$, allowing us to expand any function $f\in L^2(\X)$ as a Fourier series
\begin{equation}
f(x) = \sum_{i\ge 1} \langle \phi_i,f\rangle_\X \phi_i(x)\,.
\end{equation}
Aflalo \etal \cite{aflalo2015optimality} have recently shown that Laplacian eigenbases are optimal for representing smooth functions on manifolds. 

%Note that since the manifold Laplacian is invariant to isometries (\eg, changes in pose), nearly-isometric shapes will have approximately the same eigenfunctions (up to sign) and eigenvalues.

%----------------------------------------------------------------
\vspace{1ex}\noindent\textbf{Functional correspondence.}
In order to compactly encode correspondences between shapes, we make use of the functional map representation introduced by Ovsjanikov~\etal~\cite{ovsjanikov12}. The key idea is to identify correspondences by a linear operator ${T: L^2(\X) \rightarrow L^2(\Y)}$, mapping functions on $\X$ to functions on $\Y$. This can be seen as a generalization of classical point-to-point matching, which is a special case where delta functions are mapped to delta functions.

The linear operator $T$ admits a matrix representation $\C=(c_{ij})$ with coefficients $c_{ji} = \langle \psi_j,T\phi_i \rangle_{\Y}$, where $\{\phi_i\}_{i\geq 1}$ and $\{\psi_j\}_{j\geq 1}$ are orthogonal bases on $L^2(\X)$ and $L^2(\Y)$ respectively, leading to the expansion:
\begin{equation}\label{eq:tf}
Tf = \sum_{ij\geq 1} \langle  \phi_i,f \rangle_{\X} {c_{ji}} \psi_j\,.
\end{equation}
%
% computed as follows. Let $\{\phi_i\}_{i\geq 1}$ and $\{\psi_i\}_{i\geq 1}$ be orthogonal bases on $L^2(\X)$ and $L^2(\Y)$ respectively, and let $f\in L^2(\X)$. Then
%
%\begin{eqnarray}
%T f &=& 
%T  \sum_{i\geq 1} \langle \phi_i,f \rangle_{\X} \phi_i 
%= \sum_{i\geq 1} \langle \phi_i,f \rangle_{\X} T \phi_i \nonumber\\\label{eq:tf}
%&=& \sum_{ij\geq 1} \langle  \phi_i,f \rangle_{\X} 
%\underbrace{\langle \psi_j,T\phi_i \rangle_{\Y}}_{c_{ji}} \psi_j\,.
%\end{eqnarray}
%
A good choice for the bases $\{\phi_i\}$, $\{\psi_j\}$ is given by the Laplacian eigenfunctions on the two shapes \cite{ovsjanikov12,aflalo2015optimality}. This choice is particularly convenient, since (by analogy with Fourier analysis) it allows to truncate the series \eqref{eq:tf} after the first $k$ coefficients -- yielding a band-limited approximation of the original map. The resulting matrix $\C$ is a $k\times k$ compact representation of a correspondence between the two shapes, where typically $k\ll n$ (here $n$ is the number of points on each shape).
%It uses far less variables than the classical representation as a $n\times n$ correspondence matrix (here $n$ is the number of points on each shape), since typically $k \ll n$.
%
%Further, if the functional map $T$ is built on top of a near-isometry, one obtains $c_{ij} = \langle T \phi_i,\psi_j\rangle_\Y \approx \pm\delta_{ij}$ since near-isometric shapes have corresponding eigenfunctions (up to sign). This results in matrix $\C$ being diagonally dominant, since $c_{ij}\approx 0$ if $i\neq j$.

Functional correspondence problems seek a solution for the matrix $\C$, given a set of corresponding functions $f_i\in L^2(\X)$ and $g_i\in L^2(\Y)$, $i=1,\dots,q$, on the two shapes. In the Fourier basis, these functions are encoded into matrices $\hat{\mathbf{F}}=(\langle \phi_i , f_j\rangle_\X)$ and $\hat{\mathbf{G}}=(\langle \psi_i , g_j\rangle_\Y)$, leading to the least-squares problem:
\begin{equation}\label{eq:ls}
\min_\C \| \C \hat{\mathbf{F}} - \hat{\mathbf{G}} \|_F^2\,.
\end{equation}
In practice, dense $q$-dimensional descriptor fields (\eg, HKS \cite{sun09} or SHOT \cite{tombari10}) on $\X$ and $\Y$ are used as the corresponding functions.

%----------------------------------------------------------------
\vspace{1ex}\noindent\textbf{Label space.}
Previous approaches at learning shape correspondence phrased the matching problem as a {\em labelling} problem \cite{rodola14,masci15,add16,boscaini2016learning,monet}. These approaches attempt to label each vertex of a given query shape $\X$ with the index of a corresponding point on some reference shape $\Z$ (usually taken from the training set), giving rise to a dense point-wise map $T_\X:\X\to\Z$. The correspondence between two queries $\X$ and $\Y$ can then be obtained via the composition $T_\Y^{-1} \circ T_\X$ \cite{rodola14}.

Given a training set $S=\{(x,\pi^*(x))\}\subset\X\times\Y$ of matches under the ground-truth map $\pi^*:\mathcal{X}\to\mathcal{Y}$, label-based approaches compute a descriptor $F_\Theta(x)$ whose optimal parameters
%
% encoding a probability distribution on $\Y$, acting as a `soft' correspondence. The optimal parameters of the descriptor
%
 are found by minimizing the {\em multinomial regression} loss:
\begin{equation}\label{eq:logistic}
%\ell_{mr} (\Theta) = -\sum_{ (x,y^*(x)) \in S } \log F_\Theta (x,y^*(x))\,,
\ell_{\mathrm{mr}}(\Theta) = - \sum_{ (x,\pi^*(x)) \in S } \langle \delta_{\pi^*(x)} , \log F_\Theta (x) \rangle_\Y \,,
\end{equation}
where $\delta_{\pi^*(x)}$ is a delta function on $\Y$ at point $\pi^*(x)$.

Such an approach essentially treats the correspondence problem as one of classification, where the aim is to approximate as closely as possible (in a statistical sense) the correct label for each point. The actual construction of the full correspondence is done {\em a posteriori} by a composition step with an intermediate reference domain, or by solving the least-squares problem \eqref{eq:ls} with the learned descriptors as data.

\vspace{1ex}\noindent\textbf{Discretization.}
In the discrete setting, shapes are represented as manifold triangular meshes with $n$ vertices (in general, different for each shape). The Laplace-Beltrami operator $\Delta$ is discretized as a symmetric $n\times n$ matrix $\mathbf{L} = \mathbf{A}^{-1} \mathbf{W}$ using a classical linear FEM scheme \cite{meyer2003:ddg}, where the {\em stiffness matrix} $\mathbf{W}$ contains the cotangent weights, and the {\em mass matrix} $\mathbf{A}$ is a diagonal matrix of vertex area elements.  The manifold inner product $\langle f,g\rangle$ is discretized as the area-weighted dot product $\mathbf{f}^\top \mathbf{A} \mathbf{g}$, where the vectors $\mathbf{f},\mathbf{g}\in\mathbb{R}^{n}$ contain the function values of $f$ and $g$ at each vertex. Note that under such discretization we have $\bm{\Phi}^\top \mathbf{A} \bm{\Phi}=\mathbf{I}$, where $\bm{\Phi}$ contains the Laplacian eigenfunctions as its columns.

\begin{figure}[t]
  \centering
\begin{overpic}
[trim=0cm 0cm 0cm 0cm,clip,width=0.7\linewidth]{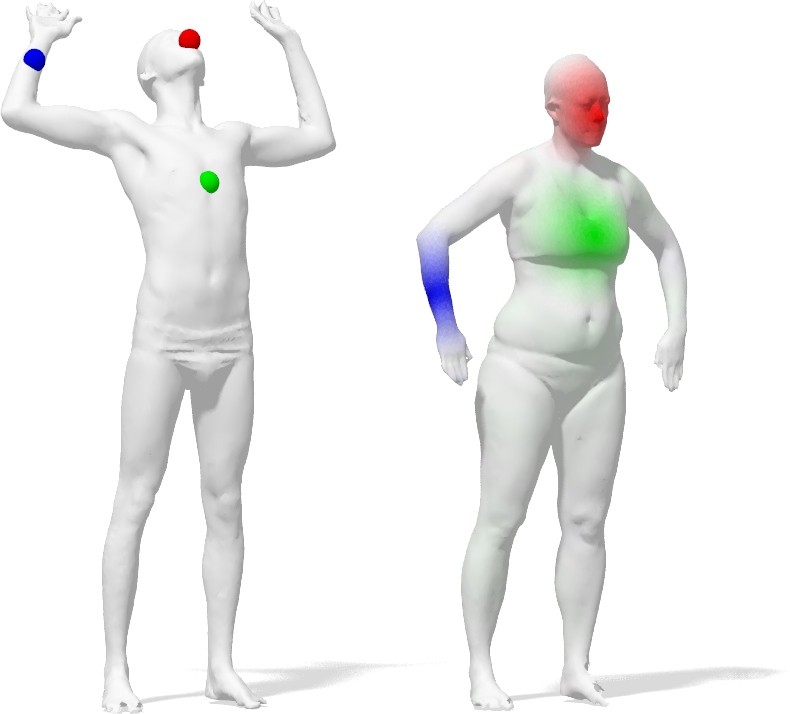}
%\put(2,-1){\footnotesize source}
\end{overpic}
  \caption{\label{fig:soft_corr}Given a source and a target shape as input, our network outputs a soft correspondence matrix whose columns can be interpreted as probability distributions over the target shape.}
\end{figure}

\begin{figure*}[t]
\centering
\begin{overpic}
		[trim=0cm 0cm 0cm 0cm,clip,width=1\linewidth]{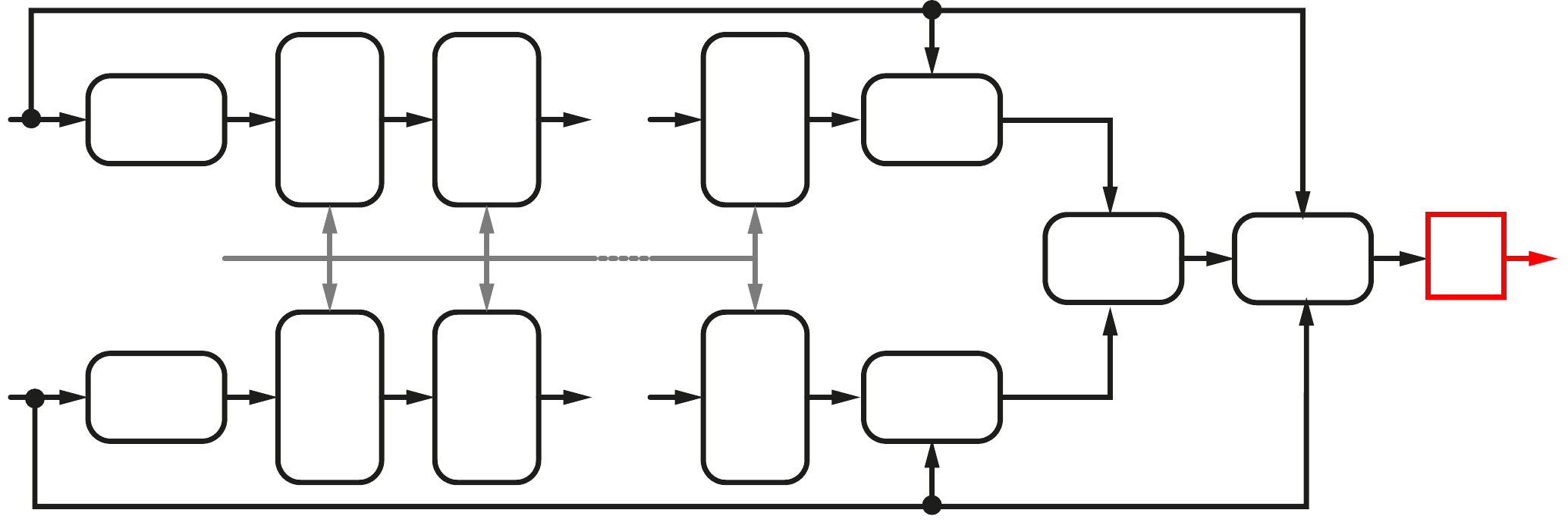}
		\put(-2,25.5){ $\mathcal{X}$}
		\put(3,30.3){ $\bb{\Phi}$}
		\put(6.8,25.4){ SHOT}
		\put(19.6,24){ \rotatebox{90}{Res 1}}
		\put(29.7,24){ \rotatebox{90}{Res 2}}
		\put(47,24){ \rotatebox{90}{Res K}}
		\put(37.8,25.6){ $\cdots$}
		\put(57,25.6){ $\langle \cdot , \cdot \rangle_\mathcal{X} $}
		\put(67.5,23){ $\hat{\mathbf{F}}$}
		\put(52.5,23){ $\bb{F}$}

		\put(10.9,16.7){ $\bb{\Theta}$}		
		\put(69,16.5){ FM}

		\put(75.7,18.5){ $\bb{C}$}
		\put(88,18.5){ $\bb{P}$}

		\put(79.5,16.6){ Softcor}
		\put(92.1,16.7){ {\color{red} $\ell_{\mathrm{F}}$ } }

		\put(3,2.5){ $\bb{\Psi}$}
		\put(-2,8){ $\mathcal{Y}$}
		\put(6.8,7.6){ SHOT}
		\put(19.6,6){ \rotatebox{90}{Res 1}}
		\put(29.7,6){ \rotatebox{90}{Res 2}}
		\put(47,6){ \rotatebox{90}{Res K}}
		\put(37.8,8){ $\cdots$}
		\put(57,7.8){ $\langle \cdot , \cdot \rangle_\mathcal{Y} $}
		\put(67.5,10){ $\hat{\mathbf{G}}$}
		\put(52.2,10){ $\bb{G}$}
\end{overpic}
\caption{\label{fig_denoiseNet}\small \textbf{FMNet architecture.} 
		Input point-wise descriptors (SHOT \cite{tombari10} in this paper) from a pair of shapes are passed through an identical sequence of operations (with shared weights), resulting in refined descriptors $\mathbf{F}, \mathbf{G}$. These, in turn, are projected onto the Laplacian eigenbases $\bm{\Phi},\bm{\Psi}$ to produce the spectral representations $\hat{\mathbf{F}}, \hat{\mathbf{G}}$. The functional map (FM) and soft correspondence (Softcor) layers, implementing Equations~\eqref{eq:ls} and \eqref{eq:soft_corr} respectively, are not parametric and are used to set up the geometrically structured loss $\ell_\mathrm{F}$ \eqref{eq:lossf}.
	}
\end{figure*}
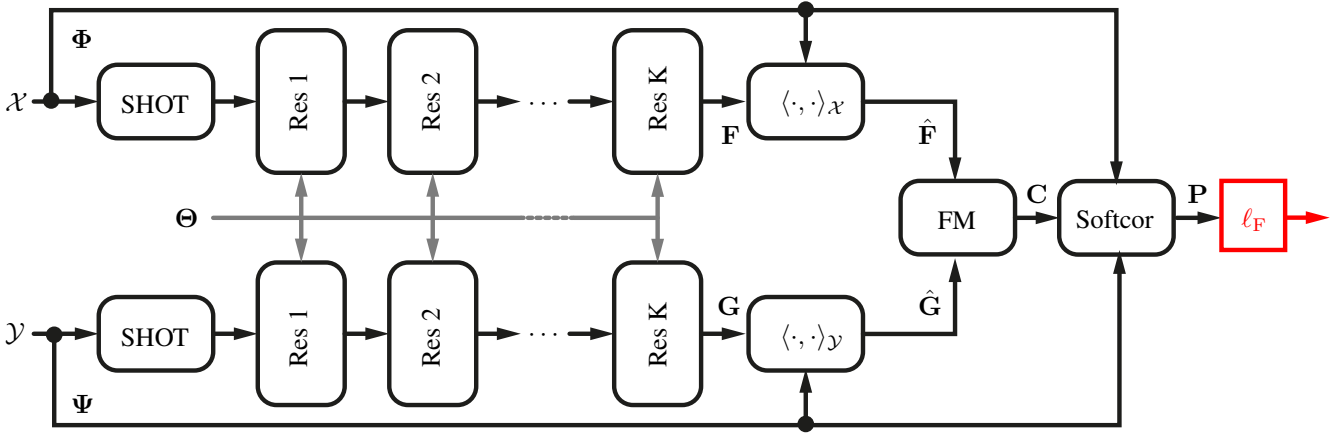

\section{Deep Functional Maps}\label{sec_method}
In this paper we propose an alternative model to the labelling approach described above. We aim at learning point-wise descriptors which, when used in a functional map pipeline such as \eqref{eq:ls}, will induce an accurate correspondence. To this end, we construct a neural network which takes as input existing, manually designed descriptors and improves upon those while satisfying a {\em geometrically} meaningful criterion. Specifically, we consider the {\em soft error loss}
\begin{align}\label{eq:lossf}
\ell_\mathrm{F} = \hspace{-0.4cm} \sum_{(x,y)\in (\mathcal{X},\mathcal{Y})} \hspace{-0.4cm} P(x,y) d_\mathcal{Y} (y, \pi^\ast (x)) = \| \mathbf{P} \circ \mathbf{D}_\mathcal{Y} \|_\mathrm{F}\,,
\end{align}
where $\mathbf{D}_\mathcal{Y}$ is the $n\times n$ matrix of geodesic distances on $\mathcal{Y}$, $\circ$ is the element-wise product, and 
\begin{equation}\label{eq:soft_corr}
\mathbf{P} = | \bm{\Psi} \C \bm{\Phi}^\top\mathbf{A}  |^\wedge
\end{equation}
is a {\em soft correspondence} matrix, which can be interpreted as the probability of point $x \in \mathcal{X}$ mapping to point $y\in\mathcal{Y}$ (see Figure~\ref{fig:soft_corr}); here, $\bm{\Phi}, \bm{\Psi}$ are matrices containing the first $k$ eigenfunctions $\{\phi_i\}$, $\{\psi_j\}$ as their columns, $|\cdot|$ acts element-wise, and $\mathbf{X}^\wedge$ is a column-wise normalization of $\mathbf{X}$. In the formula above, the $k\times k$ matrix $\mathbf{C}$ represents a functional map obtained as the least-squares solution to \eqref{eq:ls} under {\em learned} descriptors $\mathbf{F},\mathbf{G}$. 

Matrix $\mathbf{P}$ represents a rank-$k$ approximation of the spatial correspondence between the two shapes, thus allowing us to interpret the soft error \eqref{eq:lossf} as a probability-weighted geodesic distance from the ground-truth. This measure, introduced in \cite{kovnatsky15} as an evaluation criterion for soft maps, endows our solutions with guarantees of mapping nearby points on $\mathcal{X}$ to nearby points on $\mathcal{Y}$. On the contrary, the classification cost \eqref{eq:logistic}, adopted by existing label-based correspondence learning approaches, considers {\em equally} correspondences that deviate from the ground-truth, no matter how far.
Further, notice that Equation~\eqref{eq:soft_corr} is asymmetric, implying that each pair of training shapes can be used twice for training (i.e., in both directions). Also note that, differently from previous approaches operating in the label space, in our setting the number of effective training examples (i.e. pairs of shapes) increases {\em quadratically} with the number of shapes in the collection. This is a significant advantage in situations with scarce training data.

We implement descriptor learning using a Siamese residual network architecture \cite{he2016deep}. To this network, we concatenate additional {\em non}-parametric layers implementing the least-squares solve \eqref{eq:ls} followed by computation of the soft correspondence according to \eqref{eq:soft_corr}. In particular, the solution to \eqref{eq:ls} is obtained in closed form as $\mathbf{C}=\hat{\mathbf{G}} \hat{\mathbf{F}}^\dagger$, where $^\dagger$ denotes the pseudo-inverse operation. The complete architecture (named ``FMNet'') is illustrated in Figure~\ref{fig_denoiseNet}.

\section{Implementation details}\label{sec:impl}

\vspace{1ex}\noindent\textbf{Data.}
For increased efficiency, we down-sample the input shapes to 15K vertices by edge contraction \cite{garland97}; in case the input mesh has a smaller amount of vertices, it is kept at full resolution. As input feature for the network we use the 352-dimensional SHOT descriptor \cite{tombari10}, computed on all vertices of the remeshed shapes. The choice of the descriptor is mainly driven by its fast computation and its local nature, making it a robust candidate in the presence of missing parts. Note that while this descriptor is {\em not}, strictly speaking, deformation invariant, it was shown to work well for the deformable setting in practice \cite{rodola16-partial,litany16}. Recent work on learning-based shape correspondence makes use of the same input feature \cite{boscaini2016learning,monet}.

\vspace{1ex}\noindent\textbf{Network.}
Our network architecture consists of $7$ fully-connected residual layers as described in \cite{he2016deep} with exponential linear units (ELUs) \cite{clevert2015fast} and no dimensionality reduction, implemented in TensorFlow\footnote{Code and data are available at \url{https://github.com/orlitany/DeepFunctionalMaps}.} \cite{abadi2015tensorflow}. Depending on the dataset, we used 20K (FAUST synthetic), 100K (FAUST real scans), and 1K (SHREC'16) training mini-batches, each containing $\sim$1K randomly chosen ground-truth matches. For the FAUST real dataset, sampling was weighted according to the vertex area elements to prevent unduly aggregation of matches to high-resolution portions of the surface. For the three datasets, we used respectively $k=120,70,100$ eigenfunctions for representing the $k\times k$ matrix $\C$ inside the network. Training was done using the ADAM optimizer \cite{DBLP:journals/corr/KingmaB14} with a learning rate of $\alpha=10^{-3}$, $\beta_1=0.9$, $\beta_2=0.999$ and $\epsilon=10^{-8}$. The average prediction runtime for a pair of FAUST models is $0.25$ seconds.

\vspace{1ex}\noindent\textbf{Upscaling.}
Given two down-sampled shapes $\tilde{\X}$ and $\tilde{\Y}$, the network predicts a $k \times k$ matrix $\tilde{\C}$ encoding the correspondence between the two. Since this matrix is expressed w.r.t. basis functions $\{\tilde{\phi}_i\}_i, \{\tilde{\psi}_j\}_j$ of the {\em low-resolution} shapes, it can not be directly used to recover a point-wise map between the full-resolution counterparts $\X$ and $\Y$. Therefore, we perform an upscaling step as follows.

Let ${\pi_\X:\tilde{\X}\to\X}$ be the injection mapping each point in $\tilde{\X}$ to the corresponding point in the full shape $\X$ (this map can be easily recovered by a simple nearest-neighbor search in $\mathbb{R}^3$), and similarly for shape $\Y$.
Further, denote by ${\tilde{T}:\tilde{\X}\to\tilde{\Y}}$ the point-to-point map recovered from $\tilde{\C}$ using the baseline recovery approach of \cite{ovsjanikov12}. A map $T:\X\supset\mathrm{Im}(\pi_\X)\to\Y$ is obtained via the composition ${T = \pi_\Y \circ \tilde{T} \circ \pi_\X^{-1}}$. However, while $\tilde{T}$ is dense in $\tilde{\X}$, the map $T$ is {\em sparse} in $\X$.
In order to map {\em each} point in $\X$ to a point in $\Y$, we construct pairs of delta functions $\delta_{x_i}:\X\to\{0,1\}$ and $\delta_{T(x_i)}:\Y\to\{0,1\}$ supported at corresponding points $(x_i,T(x_i))$ for $i=1,\dots,|\tilde{\X}|$; note that we have as many corresponding pairs as the number of vertices in the low-resolution shape $\tilde{\X}$. We use these corresponding functions to define the minimization problem:
\begin{equation}\label{eq:upscale}
\C^*=\arg\min_\C \| \C \hat{\mathbf{F}} - \hat{\mathbf{G}} \|_{2,1}\,,
\end{equation}
where $\hat{\mathbf{F}}=(\langle \phi_i , \delta_{x_j}\rangle_\X)$ and $\hat{\mathbf{G}}=(\langle \psi_i , \delta_{T(x_j)}\rangle_\Y)$ contain the Fourier coefficients (in the full-resolution basis) of the corresponding delta functions, and the $\ell_{2,1}$-norm allows to discard potential mismatches in the data\footnote{The matrix norm $\|\mathbf{X}\|_{2,1}$ is defined as the sum of the $\ell_2$ norms of the columns of $\mathbf{X}$.}. Problem~\eqref{eq:upscale} is non-smooth and convex, and can be solved globally using ADMM-like techniques. A dense point-to-point map between $\X$ and $\Y$ is finally recovered from the optimal functional map $\C^*$ by the nearest-neighbor approach of \cite{ovsjanikov12}.

\begin{figure}[bt]
  \centering
  \input{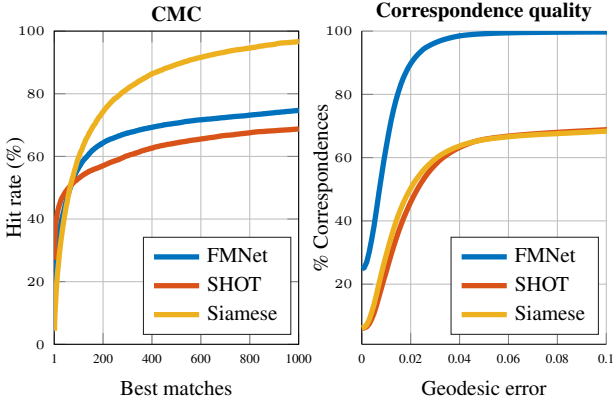}\hspace{-0.3cm}
\definecolor{mycolor1}{rgb}{0.00000,0.44700,0.74100}%
\definecolor{mycolor2}{rgb}{0.85000,0.32500,0.09800}%
\definecolor{mycolor3}{rgb}{0.92900,0.69400,0.12500}%
\begin{tikzpicture}

\begin{axis}[%
width=0.39\linewidth,
height=0.5\linewidth,
scale only axis,
xmin=0,
xmax=0.1,
xlabel={\footnotesize Geodesic error},
every x tick label/.append style={font=\color{black}, font=\tiny},
every y tick label/.append style={font=\color{black}, font=\tiny},
xmajorgrids,
xlabel near ticks,
xtick={0,0.02,0.04,0.06,0.08,0.1},
xticklabels={0,0.02,0.04,0.06,0.08,0.1},
ymin=1,
ymax=100,
title style={font=\bfseries,at={(0.5,0.95)}},
title={\footnotesize Correspondence quality},
ylabel style={at={(0.22,0.5)}},
ylabel={\footnotesize \% Correspondences},
ymajorgrids,
axis background/.style={fill=white},
axis x line*=bottom,
axis y line*=left,
legend style={at={(0.97,0.03)},anchor=south east,legend cell align=left,align=left,draw=white!15!black}
]

\addplot [color=mycolor1,solid,line width=2.0pt]
  table[row sep=crcr]{%
0	24.8243831640058\\
0.001	25.2296081277213\\
0.002	26.9184325108853\\
0.003	30.0095791001451\\
0.004	33.9291727140784\\
0.005	38.3146589259797\\
0.006	43.4699564586357\\
0.007	48.8400580551524\\
0.008	53.8934687953556\\
0.009	58.8716981132075\\
0.01	63.388969521045\\
0.011	67.7033381712627\\
0.012	71.7166908563135\\
0.013	75.1634252539913\\
0.014	78.1947750362845\\
0.015	80.8229317851959\\
0.016	83.1918722786647\\
0.017	85.2139332365747\\
0.018	86.877503628447\\
0.019	88.4296081277213\\
0.02	89.6809869375907\\
0.021	90.8185776487663\\
0.022	91.799709724238\\
0.023	92.6583454281567\\
0.024	93.4208998548621\\
0.025	94.1361393323657\\
0.026	94.7132075471698\\
0.027	95.1384615384616\\
0.028	95.5613933236575\\
0.029	95.9404934687954\\
0.03	96.2920174165457\\
0.031	96.6165457184325\\
0.032	96.910885341074\\
0.033	97.1805515239478\\
0.034	97.4145137880987\\
0.035	97.6249637155298\\
0.036	97.8269956458635\\
0.037	98.0142235123368\\
0.038	98.1950653120464\\
0.039	98.3492017416546\\
0.04	98.4830188679245\\
0.041	98.5851959361393\\
0.042	98.6786647314949\\
0.043	98.7593613933237\\
0.044	98.8403483309144\\
0.045	98.9149492017416\\
0.046	98.9747460087083\\
0.047	99.0296081277213\\
0.048	99.0789550072569\\
0.049	99.1268505079826\\
0.05	99.1660377358491\\
0.051	99.2029027576197\\
0.052	99.2397677793904\\
0.053	99.2728592162555\\
0.054	99.310595065312\\
0.055	99.3396226415094\\
0.056	99.3645863570392\\
0.057	99.388388969521\\
0.058	99.4171262699565\\
0.059	99.4351233671989\\
0.06	99.4478955007257\\
0.061	99.4647314949202\\
0.062	99.4853410740203\\
0.063	99.5027576197387\\
0.064	99.5201741654572\\
0.065	99.5329462989841\\
0.066	99.5436865021771\\
0.067	99.5521044992743\\
0.068	99.5619738751814\\
0.069	99.5689404934688\\
0.07	99.5767779390421\\
0.071	99.5820029027576\\
0.072	99.5927431059506\\
0.073	99.6002902757619\\
0.074	99.6092888243831\\
0.075	99.6162554426705\\
0.076	99.6229317851959\\
0.077	99.6281567489114\\
0.078	99.6403483309144\\
0.079	99.6432510885341\\
0.08	99.6478955007256\\
0.081	99.6545718432511\\
0.082	99.6618287373004\\
0.083	99.6647314949202\\
0.084	99.6696661828737\\
0.085	99.6754716981132\\
0.086	99.6806966618287\\
0.087	99.6859216255442\\
0.088	99.6873730043541\\
0.089	99.6902757619739\\
0.09	99.6925979680697\\
0.091	99.6952104499274\\
0.092	99.6984034833091\\
0.093	99.699564586357\\
0.094	99.7033381712627\\
0.095	99.7071117561683\\
0.096	99.7085631349782\\
0.097	99.7097242380261\\
0.098	99.710595065312\\
0.099	99.711175616836\\
0.1	99.7123367198839\\
0.101	99.7129172714078\\
};
\addlegendentry{\footnotesize FMNet};

\addplot [color=mycolor2,solid,line width=2.0pt]
  table[row sep=crcr]{%
0	6.08156748911466\\
0.001	6.15965166908563\\
0.002	6.51756168359942\\
0.003	7.39361393323657\\
0.004	8.89956458635704\\
0.005	10.7947750362845\\
0.006	13.2531204644412\\
0.007	15.9895500725689\\
0.008	18.6896952104499\\
0.009	21.4029027576197\\
0.01	23.967198838897\\
0.011	26.6943396226415\\
0.012	29.3326560232221\\
0.013	31.8452830188679\\
0.014	34.2505079825835\\
0.015	36.5236574746009\\
0.016	38.6499274310595\\
0.017	40.6746008708273\\
0.018	42.4795355587808\\
0.019	44.2597968069666\\
0.02	45.9431059506531\\
0.021	47.5164005805515\\
0.022	48.966908563135\\
0.023	50.3747460087083\\
0.024	51.6673439767779\\
0.025	52.8513788098694\\
0.026	53.9994194484761\\
0.027	55.0174165457184\\
0.028	55.9927431059506\\
0.029	56.8812772133527\\
0.03	57.6847605224964\\
0.031	58.489114658926\\
0.032	59.2171262699565\\
0.033	59.844412191582\\
0.034	60.4415094339623\\
0.035	60.9625544267054\\
0.036	61.488824383164\\
0.037	61.9706821480406\\
0.038	62.4037735849057\\
0.039	62.8229317851959\\
0.04	63.2063860667634\\
0.041	63.5587808417997\\
0.042	63.8920174165457\\
0.043	64.2043541364296\\
0.044	64.5039187227867\\
0.045	64.7851959361394\\
0.046	65.0470246734398\\
0.047	65.2597968069666\\
0.048	65.4496371552975\\
0.049	65.6142235123367\\
0.05	65.7703918722787\\
0.051	65.9079825834543\\
0.052	66.0281567489115\\
0.053	66.1451378809869\\
0.054	66.2568940493469\\
0.055	66.3532656023222\\
0.056	66.4510885341074\\
0.057	66.5439767779391\\
0.058	66.6249637155298\\
0.059	66.7155297532656\\
0.06	66.8081277213353\\
0.061	66.888534107402\\
0.062	66.9660377358491\\
0.063	67.044702467344\\
0.064	67.1123367198839\\
0.065	67.1727140783744\\
0.066	67.2435413642961\\
0.067	67.300725689405\\
0.068	67.3622641509434\\
0.069	67.4243831640058\\
0.07	67.4824383164006\\
0.071	67.5410740203193\\
0.072	67.5994194484761\\
0.073	67.6502177068215\\
0.074	67.7030478955007\\
0.075	67.7523947750363\\
0.076	67.8034833091437\\
0.077	67.8484760522496\\
0.078	67.8966618287373\\
0.079	67.9402031930334\\
0.08	67.9866473149492\\
0.081	68.0365747460087\\
0.082	68.08824383164\\
0.083	68.13381712627\\
0.084	68.1788098693759\\
0.085	68.2203193033381\\
0.086	68.2629898403483\\
0.087	68.3062409288825\\
0.088	68.3471698113207\\
0.089	68.3863570391872\\
0.09	68.4203193033381\\
0.091	68.4592162554427\\
0.092	68.5042089985486\\
0.093	68.5436865021771\\
0.094	68.5788098693759\\
0.095	68.6203193033382\\
0.096	68.655442670537\\
0.097	68.7013062409289\\
0.098	68.7474600870827\\
0.099	68.7878084179971\\
0.1	68.8272859216255\\
0.101	68.8699564586357\\
};
\addlegendentry{\footnotesize SHOT};

\addplot [color=mycolor3,solid,line width=2.0pt]
  table[row sep=crcr]{%
0	6.21654571843251\\
0.001	6.39651669085631\\
0.002	7.15471698113207\\
0.003	8.68359941944848\\
0.004	10.9759071117562\\
0.005	13.5436865021771\\
0.006	16.6534107402032\\
0.007	20.0510885341074\\
0.008	23.2772133526851\\
0.009	26.3564586357039\\
0.01	29.2525399129173\\
0.011	32.0856313497823\\
0.012	34.8197387518142\\
0.013	37.3753265602322\\
0.014	39.7965166908563\\
0.015	41.9352685050798\\
0.016	43.888824383164\\
0.017	45.7404934687954\\
0.018	47.3422351233672\\
0.019	48.8539912917271\\
0.02	50.2693759071118\\
0.021	51.6293178519594\\
0.022	52.7872278664732\\
0.023	53.8626995645863\\
0.024	54.8568940493469\\
0.025	55.8345428156749\\
0.026	56.6986937590711\\
0.027	57.4626995645864\\
0.028	58.1941944847605\\
0.029	58.9204644412191\\
0.03	59.5384615384615\\
0.031	60.155297532656\\
0.032	60.6960812772133\\
0.033	61.1712626995646\\
0.034	61.611320754717\\
0.035	62.0246734397678\\
0.036	62.3939042089986\\
0.037	62.7599419448476\\
0.038	63.0894049346879\\
0.039	63.3930333817126\\
0.04	63.6859216255443\\
0.041	63.9683599419449\\
0.042	64.2124818577649\\
0.043	64.4484760522496\\
0.044	64.6522496371553\\
0.045	64.8595065312047\\
0.046	65.0621190130624\\
0.047	65.222641509434\\
0.048	65.3869375907112\\
0.049	65.5378809869376\\
0.05	65.6760522496372\\
0.051	65.7941944847605\\
0.052	65.9100145137881\\
0.053	66.0188679245283\\
0.054	66.1190130624093\\
0.055	66.2069666182874\\
0.056	66.2870827285922\\
0.057	66.3634252539913\\
0.058	66.4499274310595\\
0.059	66.5309143686502\\
0.06	66.6095791001451\\
0.061	66.6809869375907\\
0.062	66.7590711175617\\
0.063	66.8214804063861\\
0.064	66.8835994194485\\
0.065	66.9419448476052\\
0.066	66.9927431059506\\
0.067	67.0403483309144\\
0.068	67.0876632801161\\
0.069	67.1373004354136\\
0.07	67.1895500725689\\
0.071	67.2481857764877\\
0.072	67.2934687953556\\
0.073	67.3312046444122\\
0.074	67.3680696661829\\
0.075	67.4026124818578\\
0.076	67.4374455732946\\
0.077	67.4731494920174\\
0.078	67.5126269956458\\
0.079	67.5506531204644\\
0.08	67.5936139332366\\
0.081	67.632510885341\\
0.082	67.6705370101596\\
0.083	67.7033381712627\\
0.084	67.7375907111756\\
0.085	67.7767779390421\\
0.086	67.8156748911466\\
0.087	67.855732946299\\
0.088	67.8952104499274\\
0.089	67.9309143686502\\
0.09	67.9677793904209\\
0.091	68.0075471698113\\
0.092	68.0534107402032\\
0.093	68.0937590711176\\
0.094	68.1361393323657\\
0.095	68.1761973875181\\
0.096	68.211030478955\\
0.097	68.2537010159652\\
0.098	68.2911465892598\\
0.099	68.3248185776488\\
0.1	68.3695210449928\\
0.101	68.4104499274311\\
};
\addlegendentry{\footnotesize Siamese};

\end{axis}
\end{tikzpicture}%
  \caption{\label{fig:siamese}Comparison between our structured prediction model (FMNet), metric learning (Siamese), and baseline SHOT in terms of CMC (left) and geodesic error (right). While the Siamese model produces better descriptors in terms of proximity (left), these do not necessarily induce a good functional correspondence (right).}
\end{figure}

\section{Results}\label{sec:results}

We performed a wide range of experiments on real and synthetic data to demonstrate the efficacy of our method. Qualitative and quantitative comparisons were carried out with respect to the state of the art on multiple recent benchmarks, encapsulating different matching scenarios.
%
% In order to provide a broad evaluation, we perform comparisons with both learning-based and axiomatic matching methods.

\vspace{1ex}\noindent\textbf{Error measure.}
We measure correspondence quality according to the Princeton benchmark protocol \cite{kim11}. Assume to be given a match $(x,y) \in \X \times \Y$, whereas the ground-truth correspondence is ${(x,y^*)}$. Then, we measure the {\em geodesic error}:
\begin{equation}
\epsilon(x) = \frac{d_\Y(y,y^*)}{  \mathrm{area}(\Y)^{1/2} }\,,
\end{equation}
having units of normalized geodesic length on $\Y$ (ideally, zero). We plot cumulative curves showing the percent of matches that have error smaller than a variable threshold.

\vspace{1ex}\noindent\textbf{Metric learning.}
As a proof of concept, we start by studying the behavior of our framework when the functional map layer is removed, and the soft error criterion \eqref{eq:soft_corr} is replaced with the {\em siamese} loss \cite{hadsell2006}:
\begin{align}\label{eq:siam}
\ell_{\text s}(\Theta) \hspace{-1pt} &= \hspace{-6pt} \sum_{x, x^+ \in S} \hspace{-3pt} \gamma  \| F_\Theta (x) - F_\Theta (x^+) \|^2_2\nonumber\\ 
&+ \hspace{-4pt} \sum_{x, x^- \in D} \hspace{-5pt} (1 - \gamma) (\mu - \| F_\Theta (x) - F_\Theta (x^-) \|_2)_+^2\,,
\end{align}
where $\gamma\in(0,1)$ is a trade-off parameter, $\mu>0$ is the margin, and $(x)_+=\max(0,x)$. Here, the sets $S,D\subset\X\times\Y$ constitute the training data consisting of knowingly similar and dissimilar pairs of points respectively. By considering this loss function, we transform our {\em structured prediction} model into a {\em metric learning} model. The learned descriptors $F_\Theta(x)$ can be subsequently plugged into~\eqref{eq:ls} to compute a correspondence; this metric learning approach was recently used in a functional map pipeline in \cite{clutter}.
For this test we use FAUST templates \cite{bogo14} as our data and SHOT~\cite{tombari10} as an input feature.

\begin{figure}[t]
  \centering
\begin{overpic}
[trim=0cm 0cm 0cm 0cm,clip,width=0.8\linewidth]{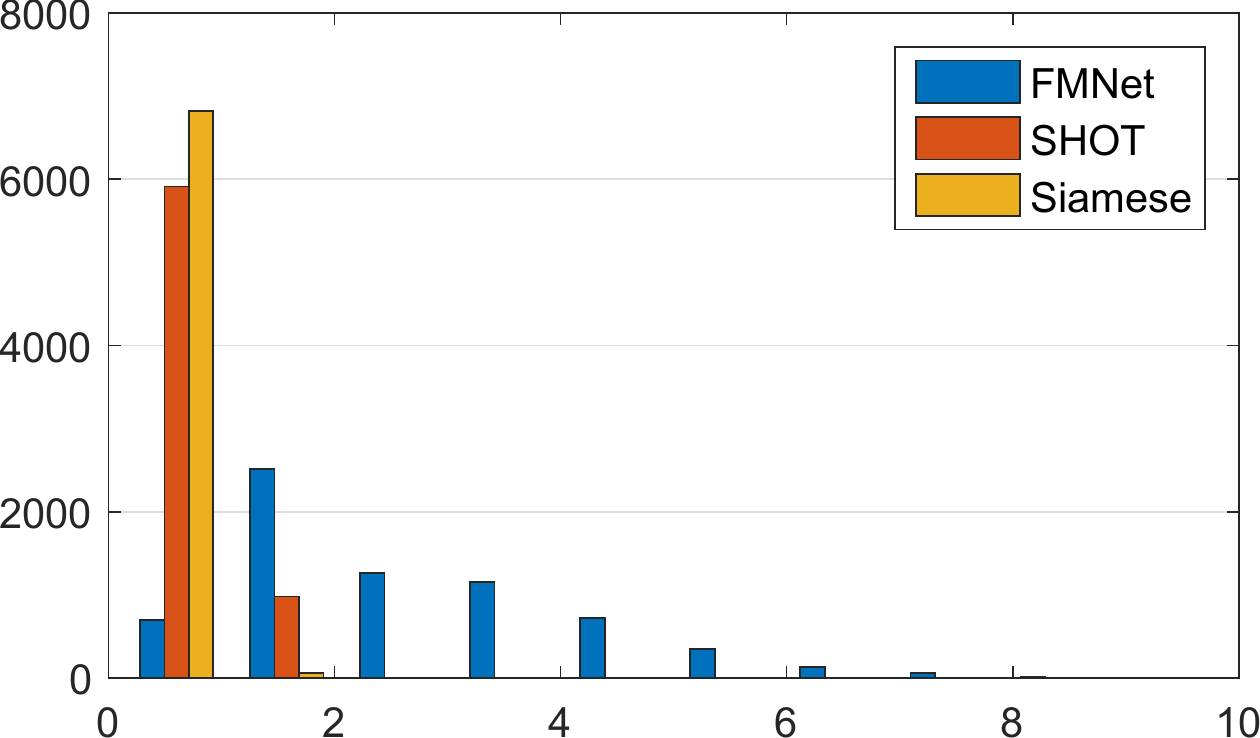}
%\put(14,10.4){\textbf{Ours}}
\end{overpic}
  \caption{\label{fig:hist}Distance distributions (in descriptor space) between correct matches. Since FMNet does not optimize a distribution criterion of this kind, it exhibits a heavy tail despite producing excellent correspondences.}
\end{figure}

From the CMC curves\footnote{{\em Cumulative error characteristic} (CMC) curves evaluate the probability ($y$-axis) of finding the correct match within the first $k$ best matches ($x$-axis), obtained as $\ell_2$-nearest neighbors in descriptor space.} of Figure~\ref{fig:siamese} (left) and the distance distributions of Figure~\ref{fig:hist} we can clearly see that the model \eqref{eq:siam} succeeds at producing descriptors that attract each other at corresponding points, while mismatches are repulsed. However, as put in evidence by Figure~\ref{fig:siamese} (right), these descriptors do not perform well when they are used for seeking a dense correspondence via \eqref{eq:ls}. Contrarily, our structured prediction model yields descriptors that are optimized for such a correspondence task, leading to a noticeable gain in accuracy.

\vspace{1ex}\noindent\textbf{Real scans.} 
We carried out experiments on real 3D acquisitions using the recent FAUST benchmark~\cite{bogo14}. The dataset consists of real scans ($\sim$200K vertices per shape) of different people in a variety of poses, acquired with a full-body 3D stereo capture system. The benchmark is divided into two parts, namely the `intra-class' (60 pairs of shapes, with each pair depicting different poses of the same subject) and the `inter-class' challenge (40 pairs of different subjects in different poses). The benchmark does not provide ground-truth correspondence for the challenge pairs, whereas the accuracy evaluation is provided by an online service. Hence, for these experiments we only compare with methods that made their results publicly available via the official ranking: the recent convex optimization approach of Chen and Koltun~\cite{chen15}, and the parametric method of Zuffi and Black~\cite{zuffi15}.

\begin{table}[bt]
\setlength{\tabcolsep}{4pt}
\centering
\begin{tabular}{|c|cccc|}
\hline
&\small\textbf{inter AE}&\small\textbf{inter WE}&\small\textbf{intra AE}&\small\textbf{intra WE}\\
\hline
Zuffi et al.~\cite{zuffi15}&3.13&6.68&1.57&5.58\\
Chen et al.~\cite{chen15}&8.30&26.80&4.86&26.57\\
FMNet&4.83&9.56&2.44&26.16\\
\hline
\end{tabular}
\vspace{1ex}
\caption{\label{tab:faust}Comparison with the state of the art in terms of average error (AE) and worst error (WE) on the FAUST challenge with real scans. The error measure is reported in cm.}
\end{table}

As training data for our approach we use the official training set of 100 shapes provided with FAUST. 
%
%As pre-processing step, each scan was remeshed to have $15K$ vertices, and its eigen-functions, Shot descriptors, and geodesic distance map (between all ot its vertices) were calculated. 
%
Since ground truth correspondences are only given between low-resolution templates registered to the scans (and not between the scans themselves), during training we augmented our data by sampling, for each template vertex, one out of 10 nearest neighbors to the real scan; this step makes the network more robust to noise in the form of small vertex displacement. %As additional augmentation while selecting random training pairs, a small subset of $1500$ corresponding vertices was sampled according to the one-ring area of the points. Thus, preventing over the sampling of denser areas like the face, and under sampling sparse ones like the stomach.

The comparison results are reported in Table~\ref{tab:faust}. Reading these results, we see that our approach considerably improves upon \cite{chen15} (around 50\%); note that while the latter method is not learning-based, it relies on a pose prior where the shapes are put into initial alignment in 3D in order to drive the optimization to a good solution.
The approach of \cite{zuffi15} obtains slightly better results than our method, but lacks in generality: it is based on a human-specific parametric model (called the `stitched puppet'), trained on a collection of 6000 meshes in different poses from motion capture data. Our model is trained on almost two orders of magnitude less data, and can be applied to any shape category (\eg, animals) as we demonstrate later in this Section.

\begin{figure}[bt]
  \centering
  \input{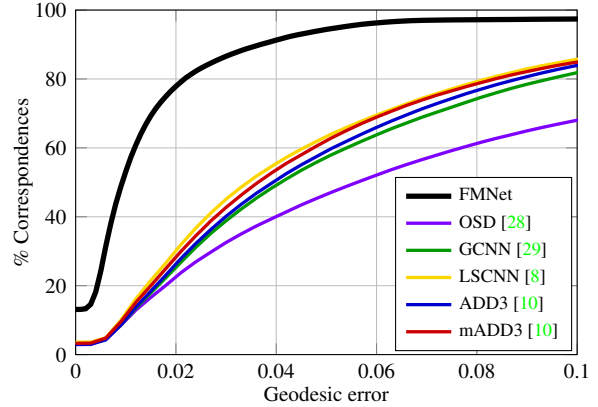}
  \caption{\label{fig:scape}Comparison with learning-based shape matching approaches on the SCAPE dataset. Our method (FMNet) was trained on FAUST data, demonstrating excellent generalization, while all other methods were trained on SCAPE.}
\end{figure}
\begin{figure}[bt]
\begin{minipage}{1.0\linewidth}
	\centering
	\input{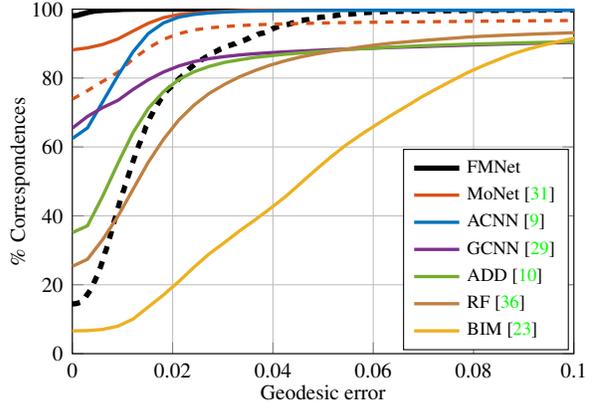}
\end{minipage}
\caption{Comparison with learning-based approaches on the FAUST humans dataset. Dashed and solid curves denote performance before and after refinement respectively. FMNet has 98\% correspondences with zero error (top left corner of the plot).}
\label{fig:monet}
\end{figure}

\vspace{1ex}\noindent\textbf{Transfer.}
We demonstrate the generalization capabilities of FMNet by performing a series of experiments on the SCAPE dataset of human shapes~\cite{anguelov05}, where our network is trained on FAUST scans data as described previously. We compare with state-of-the-art learning-based approaches for deformable shape correspondence, namely optimal spectral descriptors (OSD) \cite{litman2014learning}, geodesic CNNs (GCNN) \cite{masci15}, localized spectral CNNs (LSCNN) \cite{WFT2015}, and two variants of anisotropic diffusion descriptors (ADD3, mADD3) \cite{add16}. With the exception of FMNet, which was trained only on FAUST data, all the above methods were trained on 60 shapes from the SCAPE dataset. The remaining 10 shapes are used for testing.
%
%, reproducing verbatim the experimental setup of \cite{add16}.
%
The results are reported in Figure~\ref{fig:scape}; note that for a fair comparison, we show the {\em raw} predicted correspondence (\ie, without post-processing) for all methods.

%The dataset consists of 71 scans of a man in different poses, resulting from a template-fitting procedure on real acquisitions of a human model. Pointwise ground-truth correspondence is available.

%
\begin{figure*}[t]
  \centering
\begin{overpic}
[trim=0cm 0cm 0cm 0cm,clip,width=1\linewidth]{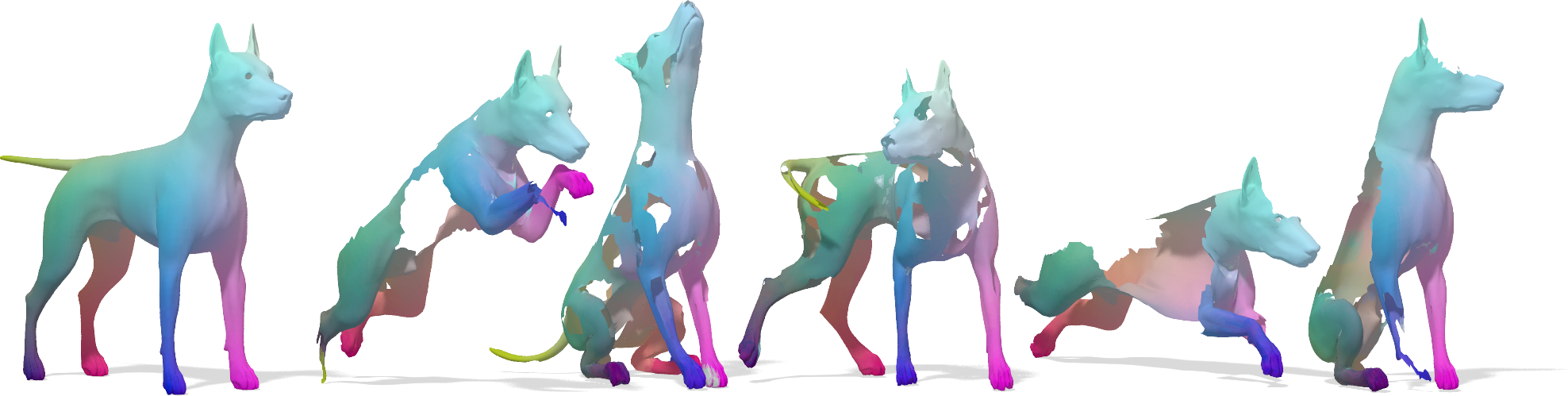}
%\put(2,-1){\footnotesize reference}
\end{overpic}
  \caption{\label{fig:shrec}Results of FMNet on the SHREC'16 Partial Correspondence benchmark. Each partial shape is matched to the full shape on the left; the color texture is transferred via the predicted correspondence.}
\end{figure*}

\vspace{1ex}\noindent\textbf{Synthetic shapes.}
For these experiments we reproduce verbatim the experimental setup of \cite{add16,boscaini2016learning,monet}: the training set consists of the first 80 shapes of the FAUST dataset; the remaining 20 shapes are used for testing. Differently from the comparisons of Table~\ref{tab:faust}, the shapes are now taken from the {\em synthetic} dataset provided with FAUST ($\sim$7K vertices per shape), for which exact ground-truth correspondence is available. We compare with the most recent state of the art in learning-based shape correspondence: random forests (RF) \cite{rodola14}, anisotropic diffusion descriptors (ADD) \cite{add16}, geodesic CNNs (GCNN) \cite{masci15}, anisotropic CNNs (ACNN) \cite{boscaini2016learning}, and the very recent MoNet model \cite{monet}. As a representative method for the family of axiomatic techniques, we additionally include blended intrinsic maps (BIM) \cite{kim11} in the comparison.

The results are reported in Figure~\ref{fig:monet}; here, all methods were post-processed with the correspondence refinement technique of \cite{pmf}. For FMNet and MoNet (the top-performing competitor) we also report curves before refinement. We note that the raw prediction of MoNet has a higher accuracy at zero. This is to be expected due to the classifier nature of this method (and all the methods in this comparison): the logistic loss \eqref{eq:logistic} aims at high point-wise accuracy, but has limited global awareness of the correspondence. Indeed, the task-driven nature of our approach induces lower accuracy at zero, but better global behavior -- note how the curve ``saturates'' at around $0.06$ while MoNet never does. As a result, FMNet significantly outperforms MoNet after refinement, producing almost ideal correspondence with zero error.

\vspace{1ex}\noindent\textbf{Partial non-human shapes.}
Our framework does not rely on any specific shape model, as it learns from the shape categories represented in the training data. In particular, it does not necessarily require the objects to be complete shapes: different forms of partiality can be tackled if adequately represented in the training set.

We demonstrate this by running our method on the recent SHREC'16 Partial Correspondence challenge~\cite{shrec16-partial}. The benchmark consists of hundreds of shapes of multiple categories with missing parts of various forms and sizes; a training set is also provided. We selected the `dog' class from the `holes' sub-challenge, being this among the hardest categories in the benchmark. The dataset is officially split into just 10 training shapes, and 26 test shapes. Qualitative examples of the obtained solutions are reported in Figure~\ref{fig:shrec}.

\section{Discussion and conclusions}\label{sec:concl}

We introduced a new neural network based method for dense shape correspondence, structured according to the functional maps framework. Building upon the recent success of end-to-end learning approaches, our network directly estimates correspondences. This is in contrast to previous descriptor learning techniques, that do not account for post processing while training. We showed this methodology to be beneficial via an evaluation on a several challenging benchmarks, comprising synthetic models, real scans with acquisition artifacts, and partiality. Being model-free, we demonstrated our method can be adapted to different shape categories such as dogs. Furthermore, we showed our method is capable of generalizing between different data-sets. 

\vspace{1ex}\noindent\textbf{Limitations.}
Laplacian eigenfunctions are inherently sensitive to topological changes. Indeed, such examples proved to be more challenging for our method. A different choice of a basis may be useful in mitigating this issue. 

\setlength{\columnsep}{5pt}
\setlength{\intextsep}{1pt}
\begin{wrapfigure}[9]{r}{0.4\linewidth}
\vspace{-1ex}
\begin{center}
		\begin{overpic}
		[trim=0cm 0cm 0cm 0cm,clip,width=0.95\linewidth]{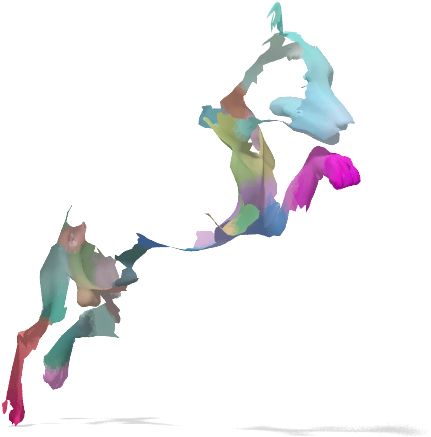}
		\end{overpic}
\end{center}
\end{wrapfigure}
As shown by recent works addressing partial correspondence using functional maps \cite{rodola16-partial}, special care should be taken when recovering matrix $\C$ from the spectral representation of the descriptors. While our method was able to recover most pairs with missing parts, it failed to recover correspondences under extreme partiality (see inset). This could be addressed by incorporating partiality priors into our structured prediction model.

%Regarding soft correspondence: 

%\vspace{1ex}\noindent\textbf{Future directions.}
%
%Future work: SVD instead of inv(C)

%There are, of course, other not the only possibility, but it has the advanteage of being differeniable and thus fit well into our end-to-end framework.

\section*{Acknowledgments}
The authors wish to thank Jonathan Masci, Federico Monti, Dan Raviv and Maks Ovsjanikov for useful discussions. OL, TR and AB are supported by the ERC grant no. 335491. ER and MB are supported by the ERC grants no. 307047 and 724228, a Google Research Faculty Award, a Radcliffe Fellowship, a Nvidia equipment grant, and a TUM-IAS Rudolf Diesel Industry Fellowship.

{\small
\bibliographystyle{ieee}
\bibliography{egbib}
}

\end{document}